\documentclass[journal]{IEEEtran}
\IEEEoverridecommandlockouts
\usepackage[english]{babel}
\usepackage[pdftex]{graphicx}
\usepackage{adjustbox}
\graphicspath{{./figures/}}
\DeclareGraphicsExtensions{.pdf,.jpg,.jpeg,.png}
\usepackage[cmex10]{amsmath}
\usepackage{bm}
\interdisplaylinepenalty=2500{}
\usepackage{physics}
\usepackage{booktabs,multirow}
\usepackage[caption=false,font=footnotesize]{subfig}

\usepackage{tikz,pgfplots}
\pgfplotsset{compat=1.3}
\usepackage{url}
\def\etal{et~al.}
\newcommand{\fref}[1]{Fig.~\ref{#1}}
\newcommand{\sfref}[2]{\fref{#1}\,\subref{#2}}
\newcommand{\tref}[1]{Table~\ref{#1}}
\newcommand{\sref}[1]{Section~\ref{#1}}

\usepackage{paralist}
\usepackage{booktabs}
\usepackage{multirow}
\usepackage{cite}

\usepackage{siunitx}

\usepackage{tikz}
\usepackage{hyperref}

\newcommand\copyrighttext{%
  \footnotesize \textcopyright 2019 IEEE. Personal use of this material is permitted.
  Permission from IEEE must be obtained for all other uses, in any current or future
  media, including reprinting/republishing this material for advertising or promotional
  purposes, creating new collective works, for resale or redistribution to servers or
  lists, or reuse of any copyrighted component of this work in other works.
  DOI: \href{https://doi.org/10.1109/TASE.2019.2918141}{10.1109/TASE.2019.2918141}}
  
\newcommand\copyrightnotice{%
\begin{tikzpicture}[remember picture,overlay]
\node[anchor=south,yshift=10pt] at (current page.south) {\fbox{\parbox{\dimexpr\textwidth-\fboxsep-\fboxrule\relax}{\copyrighttext}}};
\end{tikzpicture}%
}

\usepackage[acronym,shortcuts]{glossaries}
\glsdisablehyper
\newacronym{cmm}{CMM}{Coordinate-Measuring Machine}
\newacronym{lrf}{LRF}{Laser Range Finder}
\newacronym{svd}{SVD}{Singular Value Decomposition}
\newacronym[longplural=Degrees of Freedom]{dof}{DoF}{Degree of Freedom}
\glsunset{dof}

\makeatletter
\newcommand\np{Note\;to\;Practitioners}
\def\NP{
  \normalfont\small%
  \bfseries\textit{\np}---\relax
  \@IEEEgobbleleadPARNLSP
}

\makeatother

\title{\LARGE
  \textbf{SCALAR: Simultaneous Calibration of 2D Laser and Robot Kinematic
          Parameters Using Planarity and Distance Constraints}}

\author{Teguh Santoso Lembono$^{1,2}$, Francisco Su\'{a}rez-Ruiz$^{2}$, and Quang-Cuong Pham$^{2,3}$%
  \thanks{$^1$~Idiap Research Institute, Switzerland $^2$~School of Mechanical and Aerospace Engineering, Nanyang Technological University, Singapore \quad $^3$~Singapore Centre for 3D Printing.}}
  
\begin{document}
\maketitle

\copyrightnotice

\thispagestyle{empty}
\pagestyle{empty}

\begin{abstract}
  In this paper, we propose SCALAR, a
  calibration method to simultaneously calibrate the kinematic
  parameters of a 6-\ac{dof} robot and the extrinsic parameters of a
  2D \ac{lrf} attached to the robot's flange. The calibration setup
  requires only a flat plate with two small holes carved on it at a
  known distance from each other, and a sharp tool-tip attached to the
  robot's flange. The calibration is formulated as a nonlinear optimization problem 
  where the laser and the tool-tip are used to provide planar and distance constraints,
  and the optimization problem is solved using Levenberg-Marquardt algorithm. 
  We demonstrate through experiments that SCALAR can reduce the
  mean and the maximum tool position error from 0.44~mm to 0.19~mm and from 1.41~mm to 0.50~mm, 
  respectively.
\end{abstract}

\begin{NP}
  Many industrial robotic applications require the robotic system to
  be calibrated in order to achieve the demanded
  accuracy. Unfortunately, existing calibration processes are often
  cumbersome or require expensive measurement systems. This paper
  presents a novel calibration method to calibrate
  simultaneously a 6-\ac{dof} industrial robot and a 2D laser scanner
  attached to the robot's flange.  The method only uses the data
  from the robot and the laser, so there is no need for external
  measurement systems. The laser scanner can also be used after the
  calibration for subsequent robot tasks such as scanning a workpiece
  to accurately determine its location. The calibration setup only
  requires a flat plane and a sharp tool-tip attached on the robot's
  flange. The proposed method is easy to deploy and is more
  cost-effective than existing calibration methods.
\end{NP}

\section{Introduction}
\label{sec:introduction}

\IEEEPARstart{T}{raditional} robotics applications, such as pick and place,
spray-painting and spot-welding, rely on the high \emph{repeatability} of
existing industrial robots and hence tend to
overlook \emph{accuracy}. However, there is an increasing number of 
applications (e.g. robotic on-demand fulfillment, robotic 3D printing \cite{Zhang2018}, etc.)  where the robot must adapt dynamically to a changing environment
and hence the robot's accuracy becomes crucial.  Consider for
instance the robotic drilling task in \cite{Suarez-Ruiz2018} where the
robot is required to drill several holes at precisely-defined
locations on a workpiece. The workpiece can vary for each
task, and the placement within the workspace may not be precisely
known. To tackle such task automatically 
the robot has to scan the workspace, determine the
location of the workpiece, and finally move to the drilling
locations, all of which have to be done accurately. The accuracy 
of such a system depends on at least
two factors: the accuracy of the robot and the accuracy of the
measurement system.

\begin{figure}[t]
  \centering
  \includegraphics[width=0.7\linewidth]{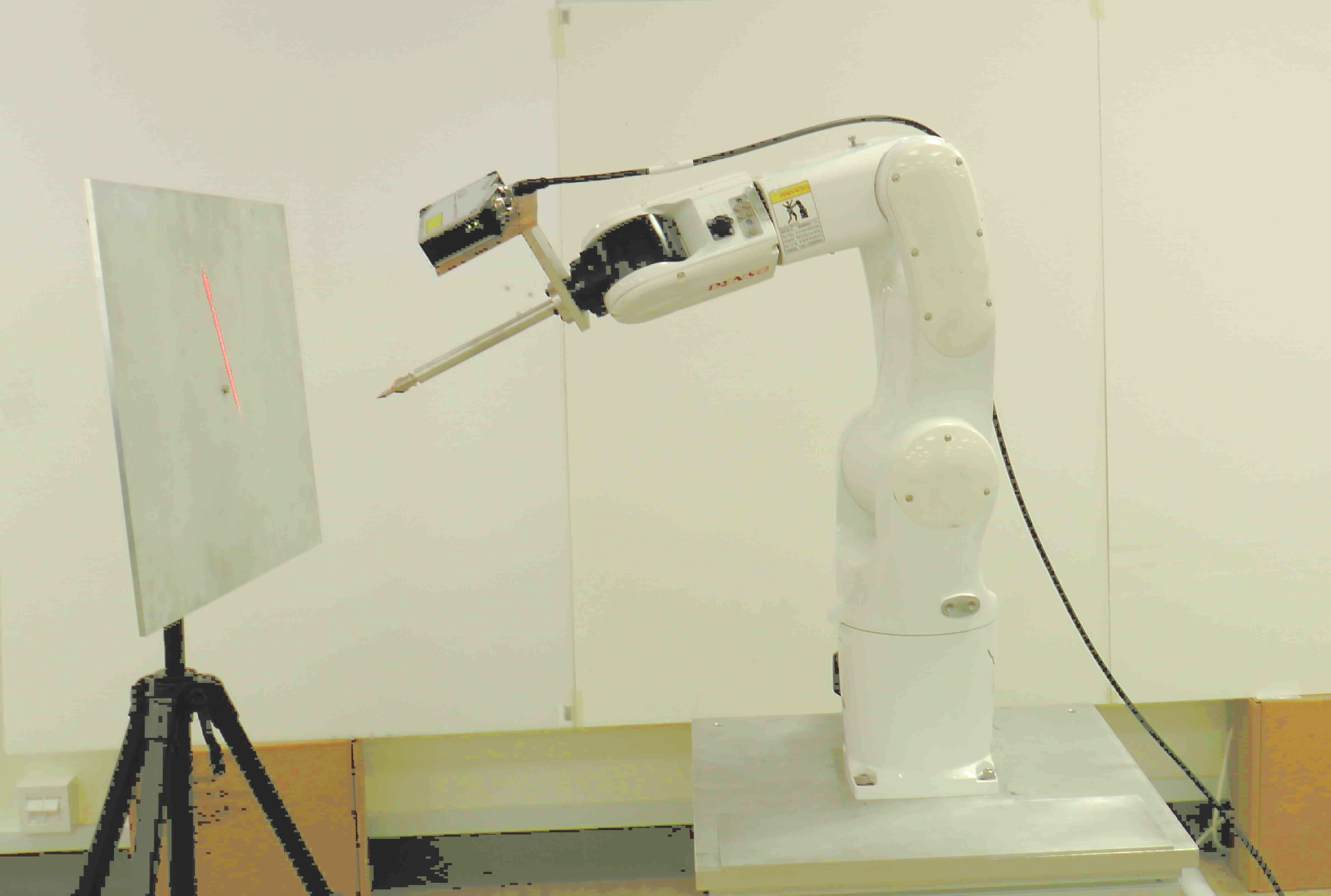}
  \caption{Calibration Setup: A 6-\ac{dof} industrial robot equipped with a 2D
  \acl{lrf} and a sharp tool. A flat plate is located within the robot's 
  workspace and the \ac{lrf}'s ray is projected onto the plate.}
  \label{fig:setup}
\end{figure}

The accuracy of the robot is determined by how close the robot's model
is to the actual kinematic parameters of the robot. \emph{Robot kinematic calibration}
is usually conducted to achieve a higher accuracy, either by an
external measurement system or by constraining the motion of the
end-effector. 

To improve the accuracy of the measurement system, \emph{intrinsic calibration} and \emph{extrinsic calibration} are often necessary. \emph{Intrinsic calibration} refines the internal parameters of the measurement devices (e.g. focal length and lens distortion for camera), while \emph{extrinsic calibration} refines the transformation between the measurement system and the robot coordinate system. In addition, the type of the measurement device also affects the accuracy. For example. a near-range laser scanner commonly used in industry for workpiece profiling can achieve a very high accuracy (on the order of \SI{}{\micro\metre}). For such high-accuracy device, the intrinsic calibration is often performed by the manufacturer, so only the extrinsic calibration is required. 

In~\cite{Lembono2018} we proposed an earlier version of SCALAR, which
we shall refer to as SCALAR$_\alpha$ in this article, to calibrate
simultaneously the kinematic parameters of a 6-\ac{dof} robot and the
extrinsic parameters of a 2D \acf{lrf} using only the information
provided by the \ac{lrf} attached to the robot's flange. The \ac{lrf}
was chosen because it costs one order of magnitude less than commonly
used measurement systems (such as Vicon or FARO Laser
Tracker). Moreover, it can be used during the subsequent robotic tasks. For example,
the robot can move the \ac{lrf} to scan the workpiece in the drilling
task \cite{Suarez-Ruiz2018} to determine the holes' positions accurately .

We reported the performance of SCALAR$_\alpha$ for a simulated
system in~\cite{Lembono2018}. During  experiments with our real setup we found that since
SCALAR$_\alpha$ calibrates \emph{simultaneously} the robot's and the laser's
parameters, each system, taken in isolation, is not properly calibrated.
That is, neither the robot's kinematic parameters, nor the laser's extrinsic parameters
converge to the good values, although the combined system is accurate. 
We did not encounter this errors during simulation but only during the experiments with the real
setup. In the experiments there are multiple sources of error that
perturb the calibration process and cause it to deviate from the
actual values.

In this paper, we present an improvement over SCALAR$_\alpha$, which will be called SCALAR in this article. 
The key idea is to introduce an additional distance constraint to the optimization problem, such that the robot's kinematic parameters converge to good values. This distance constraint is obtained by 
moving a tooltip attached to the robot's end-effector through a known distance, and recording the robot's joint values at the beginning and the end of the movement.

The remainder of the paper is as follows. In \sref{sec:related} we discuss
existing approaches for calibrating robot's kinematic
parameters and \ac{lrf}'s extrinsic parameters. In \sref{sec:method}, SCALAR is explained
in detail. The experimental results are presented in
\sref{sec:experiment_result} to verify SCALAR, and finally we conclude with a
few remarks in \sref{sec:conclusions}.

\section{Related works}
\label{sec:related}

\subsection{Calibration of robot's kinematic parameters}
\label{sec:kine_calib}
Existing robot calibration procedures can be
divided in two categories: unconstrained and constrained calibration. In unconstrained calibration, an external measurement system is required to precisely determine the pose of the robot's end-effector. Some of the examples of unconstrained calibration works include Ye~\etal~\cite{Ye2006}, Ginani and Mota \cite{Ginani2011}, Nubiola and Bonev \cite{Nubiola2013}, and Wu~\etal~\cite{Wu2017}, who use measurement systems such as Faro Laser Tracker, Romer Measurement Arm, and SMX Laser Tracker to calibrate the robot. 
There are two limitations for this type of calibration
methods: the calibration setup is cumbersome and the external
measurement systems are very expensive. 

As alternative, several researchers have focused on developing calibration methods that rely on the
sensors available in the robotic systems combined with poses restricted by
pre-engineered constraints, i.e. constrained calibration. The constraints can be in the form of point
constraints~\cite{Gatla2007}, surface constraints,
-- in~\cite{Chiu2003} the authors manufactured a cylinder and used its curved
surface as constraint -- or, planar
constraints~\cite{Zhong1995,Zhong1996,Ikits1997,Zhuang1999,Zilong2006,Hage2011,Joubair2015}.

In~\cite{Wang2009}, Wang~\etal~use point constraint in the form of a ball with a known radius, and the robot moves the laser attached to it to measure the center of the ball. In~\cite{Liu2009} and ~\cite{Liu2011low}, Liu~\etal~obtain the point constraint by using a laser pointer and a Position-Sensitive Detector (PSD), but they only calibrate the joint offset instead of the whole robot kinematic parameters. 

Ikits and Hollerbach~\cite{Ikits1997} propose a kinematic calibration method
using a planar constraint via a touch probe. While the approach is promising, they also report that some of the parameters are hardly observable when the measurements are noisy or when the model is incomplete. In~\cite{Zhuang1999}, Zhuang~\etal~investigate robot calibration with planar
constraints, in particular the observability conditions of the robot's kinematic
parameters. They prove that a single-plane constraint is insufficient for
calibrating a robot's kinematics, and a minimum of three planar
constraints are necessary. 

Joubair and Bonev~\cite{Joubair2015} calibrate both the kinematic and
non-kinematic (stiffness) parameters of a FANUC LR Mate 200iC industrial robot
using planar constraints in the form of a high precision 9-inches granite
cube. The robot is equipped with an MP250 Renishaw touch probe, which is then
moved to touch four planes of the granite cube. The granite cube's face is flat
to within $0.002$~mm. In another work~\cite{Joubair2015kinematic}, Joubair and Bonev use distance and sphere constraints in the form of a triangular plate with three 2-inch spheres separated at known distance. 

In \cite{Choi2018}, Choi \etal~calibrate the extrinsic parameters of a multi-camera system and the DH parameters of a pan-tilt unit (regarded as a 2-\ac{dof} robot manipulator) simultaneously. One of the camera is attached to the pan-tilt unit, and a fiducial marker is used to provide the reference frame for all the cameras. In their work, however, the focus is more about calibrating the camera system instead of the manipulator, and the manipulator used only has 2-\ac{dof}. In \cite{ArunDas2018}, the work is extended to a 5-\ac{dof} robot manipulator. 

\subsection{Calibration of extrinsic 2D \ac{lrf} parameters}
\label{sec:laser_calib}

Extrinsic calibration of an \ac{lrf} consists of finding the correct homogeneous
transformation from the robot coordinate frame to the laser coordinate frame.
Most of the works on extrinsic calibration of an \ac{lrf} involves a camera,
since both sensors are often used together. The works in this field are largely
based on Zhang and Pless' work~\cite{Zhang2004}. They propose a method to
calibrate both a camera and an \ac{lrf} using a planar checkerboard pattern. The checkerboard 
pattern provides the planar constraints for an optimization problem where both camera and laser
parameters are optimized simultaneously using
Levenberg-Marquardt algorithm. Unnikrishnan and Hebert~\cite{Unnikrishnan2005} use the same setup
as~\cite{Zhang2004}, but they do not optimize the camera parameter
simultaneously. Wenyu~\etal~\cite{Chen2018} propose a noise tolerant algorithm using a planar disk with arbitrary orientation to calibrate the laser sensor. Li~\etal~\cite{Li2017} calibrate a 3D laser sensor using a calibration sphere. 

\subsection{Novelty of the proposed method}
\label{sec:novelty}

SCALAR can be seen as a combination of the algorithm for extrinsic calibration
of an \ac{lrf}~\cite{Zhang2004} and the algorithm for calibration of robot's
kinematic parameters using planar constraints~\cite{Joubair2015}. In \cite{Lembono2018}~we described the advantages of SCALAR over the existing methods: a) The parameters of the robot and of the laser are calibrated simultaneously, b) No expensive external measurement system is required, c) The calibration plate can be easily manufactured, and d) Calibration poses can be distributed globally in the robot's workspace. Unlike our previous work in SCALAR$_\alpha$~\cite{Lembono2018}, SCALAR uses only two planar constraints and an additional distance constraint, and the resulting robot's kinematic parameters are more accurate. 
\section{Method}
\label{sec:method}

The calibration setup is depicted in \fref{fig:setup}, where a flat plate is
placed within the reachable robot workspace. There are two small holes
on the plate separated by a known distance $D$. An \ac{lrf} and a sharp tool-tip
are attached to the robot flange. The calibration procedure can be divided into
four steps:

\begin{enumerate}
\item The plate is moved to two ($k=2$) different locations. For each
location, the robot is moved to $N$ poses such that the \ac{lrf}'s ray is
directed to the respective plate. One reading from the 2D \ac{lrf} ray contains
hundreds of data points, so $M$ data points are selected randomly for each pose (after removing the outliers by standard line fitting with RANSAC),
and the robot's joint angles are recorded.

\item The plate is moved to $L=15$ additional locations. For each location, the
robot is moved such that the tool-tip touches the pair of holes on the plate
consecutively without changing the end-effector orientation. The tool-tip
orientation may vary between different plate locations, but at any particular location the
robot has to maintain the same orientation between the pair of holes.

\item A tool calibration is performed so that the position of the tool-tip with
respect to the robot's flange is known.

\item Finally, the Levenberg-Marquardt nonlinear optimization algorithm is used
to calibrate the all the parameters (robot kinematic and \ac{lrf} extrinsic
parameters) based on the collected data.

\end{enumerate}

The calibration algorithm remains largely the same as SCALAR$_\alpha$~\cite{Lembono2018},
except with the addition of the distance constraint and the reduction of the plane locations
from three to two locations. Hence, we will focus on the part where SCALAR differs from SCALAR$_\alpha$. 

The calibration algorithm can be described as follows. First, the initial
estimate of the \ac{lrf}'s extrinsic parameters are obtained using the linear
least-squares method with the data from one of the plates's location in Step~1.
This is based on the algorithm in~\cite{Zhang2004} and presented
in detail in ~\cite{Lembono2018}. Next, the robot kinematic parameters
and the \ac{lrf}'s extrinsic parameters are optimized simultaneously to satisfy
the planar and distance constraints using Levenberg-Marquardt nonlinear
optimization method. The optimization method requires the tool-tip coordinates
w.r.t. the robot's flange, so the next section describes the tool calibration to
obtain the tool-tip coordinates accurately. Finally, we present how \ac{svd} can
be used to determine the identifiable calibration parameters.

\subsection{Optimizing the \ac{lrf} Extrinsic and Robot's Kinematic Parameters}
\label{sec:second_step}

In SCALAR$_\alpha$~\cite{Lembono2018}, the optimization step uses the laser data from
three planes to optimize the extrinsic parameters of the \ac{lrf}, the robot's
kinematic parameters and the plane parameters -- three plane equations, one
for each plate location. The objective function in SCALAR$_\alpha$ is described as
follows,
\begin{equation}
  \label{eq:SCALAR}
  f (\vb*{\Phi}) =  \sum_{k=1}^{3} \sum_{j=1}^{N} \sum_{i=1}^{M} ({{^B}\vb*{n}_k}^T \; {^B}\vb*{p}_{ji} - {^B}l_k)^2
\end{equation}

The parameters $\vb*{\Phi}$ consist of the following:

\begin{itemize}

\item Robot's kinematic parameters. We use \emph{DH} parameters
$[a_i \;, \alpha_i \;,\theta_i \;,d_i], i=1, 2, \cdots ,6$ to
represent the robot's kinematics.

\item \ac{lrf}'s extrinsic parameters. We use the axis-angle representation 
for the rotation part $[r_x \quad r_y \quad
r_z \quad r_{\theta}]$, and $[p_x \quad p_y\quad p_z]$ for the position part.

\item Plane parameters. Each plane can be described by an unit vector
$[{^B}n_{k,x}\quad {^B}n_{k,y}\quad {^B}n_{k,z}]$ normal to the plate and its
perpendicular distance from the robot base's coordinate system origin
${^B}l_{k}$.

\end{itemize}

However, based on experimental results with a real robotic system, we found that using
the above objective function \eqref{eq:SCALAR} that optimizes the \emph{combined} robot
kinematic parameters and the \ac{lrf} extrinsic parameters \emph{based only on
the planar constraints} has one important issue. The \emph{combined} parameters indeed show good
performance minimizing the errors of the data points w.r.t. the planar
constraints -- they give us an accurate prediction of ${^B}\vb*{T}_{L}$ -- but
often fail to estimate accurately the actual end-effector pose,
${^B}\vb*{T}_{E}$, and the \ac{lrf} extrinsic parameters ${^E}\vb*{T}_{L}$. In
summary, the \emph{combined} parameters accurately estimate the laser frame
w.r.t. the base frame, ${^B}\vb*{T}_{L}$, but not necessarily the robot's
end-effector frame w.r.t the base frame, ${^B}\vb*{T}_{E}$.

\begin{figure}[t]
  \centering
  \subfloat[]{\includegraphics[width=0.49\linewidth]{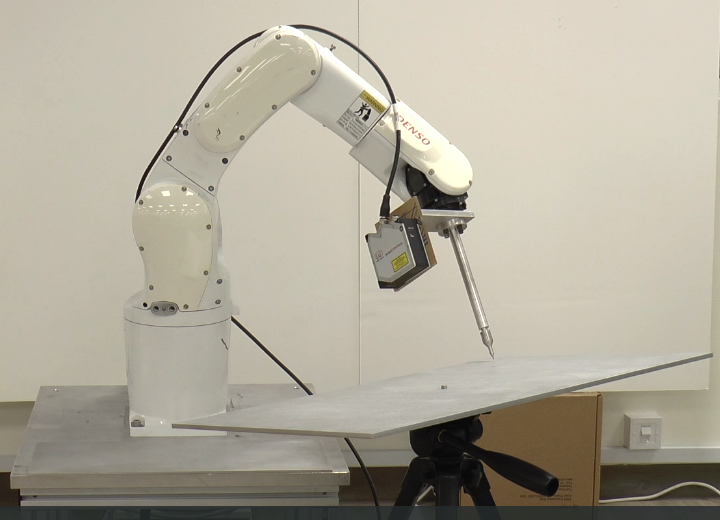}%
        \label{subfig:constraints-a}} \hfill
  \subfloat[]{\includegraphics[width=0.49\linewidth]{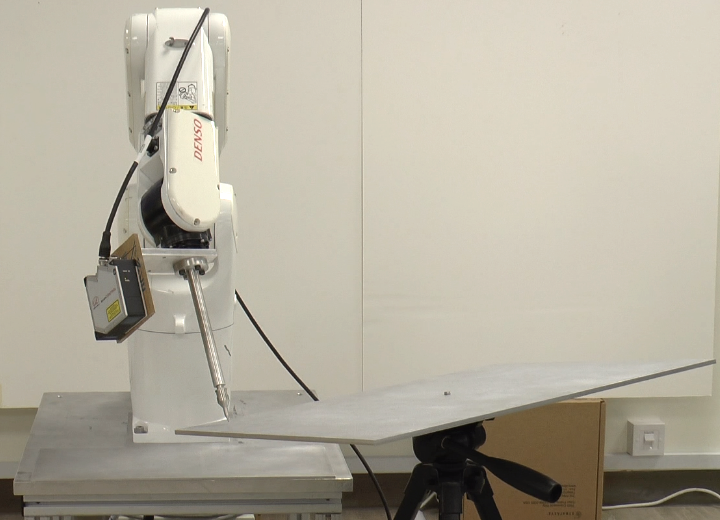}%
        \label{subfig:constraints-b}}
  \caption{Distance constraints by moving the tool-tip linearly through a known distance.}
  \label{fig:constraints}
\end{figure}

To account for this problem, we need to add more constraints that only affect
one of the two parameters sets -- either the robot or the laser parameters. To
achieve this, we move the robot's end-effector between two points that have a
known distance $D$ and use this as a \emph{distance constraint} during the calibration
optimization. The plate has a pair of holes separated at distance
$D$, and a sharp tool-tip is attached on the robot's flange. The robot is moved
such that the tool-tip touches the first hole, see
\sfref{fig:constraints}{subfig:constraints-a}, and the robot's joint angles are
recorded. At this configuration, the tool-tip position can be computed as 
${^B}\vb*{t}_{l1} = {^B}\vb*{T}_{E,l1} \; {^E}\vb*{t}$,
where ${^B}\vb*{t}_{l1}$ and ${^E}\vb*{t}$ are the position of the tool-tip in
the robot base and end-effector frame respectively, $l$ is the location index of
the plane, and the subscript~$1$ refers to the first hole. The robot is then
moved to touch the second hole, see
\sfref{fig:constraints}{subfig:constraints-b}, while maintaining the same
orientation, and the joint angles are recorded. The tool-tip position at this second
configuration can be computed as ${^B}\vb*{t}_{l2} = {^B}\vb*{T}_{E,l2} \; {^E}\vb*{t}$.

Since the distance between the two is known as $D$, then the parameters $\Phi$
have to satisfy the following constraint:
\begin{equation}
  \label{eq:additional_constraint}
 f_t (\vb*{\Phi}) =  \sum_{l=1}^{L} (|{^B}\vb*{t}_{l2} - {^B}\vb*{t}_{l1}| - D)^2 = 0 .
\end{equation}
Note that only the robot's kinematic parameters in $\vb*{\Phi}$ are affected by
the additional constraint, while the laser parameters in $\vb*{\Phi}$ are not
affected at all. This constraint is added as an additional term to the objective function, so the
objective function in SCALAR is now defined as
\begin{multline}
  \label{eq:SCALAR+}
  f (\vb*{\Phi}) =  \sum_{k=1}^{2} \sum_{j=1}^{N} \sum_{i=1}^{M} ({{^B}\vb*{n}_k}^T \; {^B}\vb*{p}_{ji} - {^B}l_k)^2 + \\  w\sum_{l=1}^{L} (|{^B}\vb*{t}_{l2} - {^B}\vb*{t}_{l1}| - D)^2
\end{multline}
where $w$ is an introduced weight to scale the contribution of the distance
constraints over the planar constraints. The value of $w$ has to be carefully chosen to balance the effect of the first and the second term in the objective function. In \sref{sec:determining_weight} the optimal value of $w$ will be determined experimentally.

$\vb*{\Phi}$ consists of 24 \emph{DH} parameters for a 6-\ac{dof} robot, 7
parameters for the laser's parameters,
and 8 parameters for the plane parameters at two locations. In total, there are
39 parameters to be optimized by minimizing the objective function
$f(\vb*{\Phi})$. The optimization problem is then solved using a
Levenberg-Marquardt nonlinear optimizer~\cite{Newville2014}.

For the \emph{unit vector} parameters $[r_x \quad r_y \quad r_z ]$ and  $[{^B}n_{k,x}
\quad {^B}n_{k,y} \quad {^B}n_{k,z}]$, the following constraints are added to
the optimization solver,
\begin{equation}
  \label{eq:10}
  {r_z} = \sqrt{1 - {r_x}^2 - {r_y}^2}
\end{equation}
\begin{equation}
  \label{eq:11}
  {^B}n_{k,z} = \sqrt{1 - {{^B}n_{k,x}}^2 - {{^B}n_{k,y}}^2}
\end{equation}

\subsection{Tool Calibration Procedure (Optional)}
\label{sec:tool-tip_calib}

The additional distance constraint in \sref{sec:second_step} contains the term
${^E}\vb*{t} = [t_x \quad t_y \quad t_z]$, which is the \emph{tool-tip
coordinates} in the end-effector frame. If the robot accurately maintains the
tool-tip orientation while touching the pair of holes, the distances travelled by
the robot's flange and the tool-tip are actually the same since they are part of
the same rigid body. In that case, an accurate value of ${^E}\vb*{t}$ is not necessary.

However, this is true under the assumption that the end-effector orientation can
be kept constant between the two hole positions. This largely
depends on the accuracy of the initial robot kinematic parameters. If these
parameters are far from the real ones, then the end-effector orientation might
actually change significantly between the two holes. In this case, the error in
${^E}\vb*{t}$ -- computed using the initial robot kinematic parameters -- will
corrupt the calibration result, hence a more accurate value of ${^E}\vb*{t}$ will be necessary.
Several methods can be used to get such accurate value, e.g. using a Coordinate-Measuring Machine to measure ${^E}\vb*{t}$ or using the information from the CAD models.

\begin{figure}[t]
  \centering
  \subfloat[]{\includegraphics[width=0.44\linewidth]{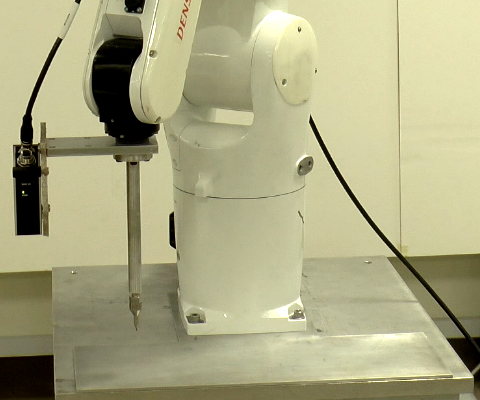}%
      \label{subfig:tool-calib-a}} \hfill
  \subfloat[]{\includegraphics[width=0.44\linewidth]{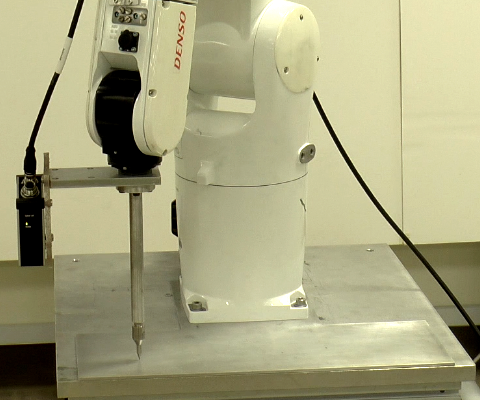}%
      \label{subfig:tool-calib-b}}
  \vspace{4mm}
  \subfloat[]{\includegraphics[width=0.45\linewidth]{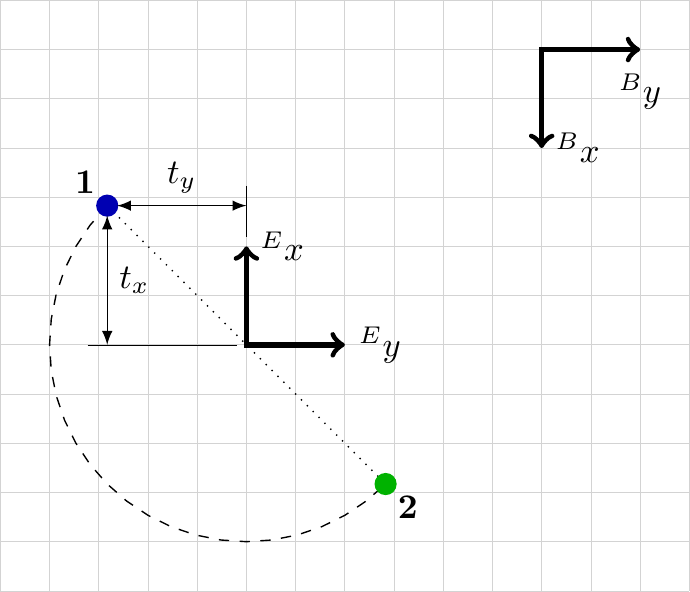}%
      \label{subfig:tool-calib-c}}
  \caption{Tool Calibration}
  \label{fig:tool-calib}
\end{figure}


In this work, we use a simple calibration method to calibrate ${^E}\vb*{t}$. A
base plate with a small diameter hole is attached to the robot base. First the tool is oriented such that z-axis of the robot's end-effector ${^E}\textbf{z}$ 
points to the opposite direction of the robot base's z-axis ${^B}\textbf{z}$ (the
vertical direction), see \sfref{fig:tool-calib}{subfig:tool-calib-a}, while ${^E}\textbf{x}$ points to the opposite of ${^B}\textbf{x}$, and ${^E}\textbf{y}$ points
to the same direction as ${^B}\textbf{y}$. Next, we move the robot such that the tool-tip
touches the hole on the base plane, see
\sfref{fig:tool-calib}{subfig:tool-calib-b}, while maintaining its orientation.
The z-coordinate of the hole is 0, because it is located on the robot base. From
the robot's kinematics, we can also get the z-coordinate of the robot's flange.
The difference between the two will give us the z-component of ${^E}\vb*{t}$,
which is $t_z$.

Next we need to find $t_x$ and $t_y$. To do that, when the robot is at the
position in \sfref{fig:tool-calib}{subfig:tool-calib-b}, the last joint of the
robot is rotated by 180 degrees. That will cause the tool-tip to move
semi-circularly in the horizontal plane. In
\sfref{fig:tool-calib}{subfig:tool-calib-c}, the center of the circle is the x-y
coordinates of the robot's flange, while the x-y coordinates of the tool-tip
will traverse the circle as the last joint is rotated. Both coordinates are seen
from the robot base frame (${^B}x$ and ${^B}y$). Initially, the tool-tip is at
point 1 (which coincides with the hole on the base plane), and the coordinate
frame of the robot flange is as indicated at the center (${^E}x$ and ${^E}y$).
At this position, $t_x$ and $t_y$ are the same as the x and y coordinates of the
point 1 in the robot's flange coordinate frame which are still unknown. As the
last joint is rotated 180 degrees, the tool-tip will move along the circle from
point 1 to point 2. By calculating the shift in x and y direction between point
1 and point 2, $t_x$ and $t_y$ can be calculated. Since point 1 coincides with
the hole, this shift can be measured by moving the tool-tip linearly without
changing orientation from point 2 to point 1 using the robot's teaching pendant,
and record the change of the robot's tool-tip position. From that we can obtain
$t_x$ and $t_y$.

In the experiment section, we will demonstrate that the accurate tool calibration 
is not necessary in SCALAR if the robot's initial \emph{DH} parameters are quite accurate. Hence, we use this tool calibration mainly
for the validation purpose. In actual calibration practice, the user can just do
a rough measurement or use the information from the CAD model to obtain the
tool-tip coordinates.

\subsection{Identifiability of the calibration parameters}
\label{sec:third_step}

In~\cite{Lembono2018} we used three plate locations for the planar constraints
based on the findings of Zhuang~\etal~\cite{Zhuang1999}. However, we realize
that two planes are actually sufficient for our method of calibration, due to
the fact that we are using a 2D laser, whereas~\cite{Zhuang1999} used a touch
probe. Unlike a touch probe which only gives a single point on the robot
end-effector for calibration, 2D laser data gives us multiple data points with
respect to the robot's end-effector, and these points change at different poses.
This result in fewer number of planes required for the calibration. Using the
identifiability analysis here, we demonstrate that two planes are indeed enough
for calibrating the same set of parameters as with three planes.

Following the approach in~\cite{Joubair2015} and~\cite{Hollerbach1996}, \ac{svd}
is applied on the identification Jacobian matrix $\vb*{J}$. Let $f_{kji}(\Phi)$
be the geometric constraint equation on the data point $i$ at the robot pose $j$
and on the plane $k$,
\begin{equation}
\label{eq:12}
 f_{kji}(\vb*{\Phi}) =  {{^B}\vb*{n}_k}^T \; {^B}\vb*{p}_{ji} - {^B}l_k = 0 \; .
\end{equation}

Then $\vb*{J}$ can be computed by differentiating \eqref{eq:12} for all the data
points $i = 1, \cdots, M$ at the robot poses $j = 1, \cdots, N$ and for all the
planes $k=1,2$, then stack them together as a matrix, $
\vb*{J} = \begin{bmatrix}
 \frac{\partial f_{111}(\vb*{\Phi})}{\partial\vb*{\Phi}} \quad
 \frac{\partial f_{112}(\vb*{\Phi})}{\partial\vb*{\Phi}} \quad
 \cdots  \quad
 \frac{\partial f_{2MN}(\vb*{\Phi})}{\partial\vb*{\Phi}} \quad
	\end{bmatrix} ^T \; .
$ We can then apply \ac{svd} to the matrix $\vb*{J} = \vb*{U}\vb*{\Sigma}\vb*{V}^T \; $.

For this identification step, the parameters $[r_z \quad {^B}n_{1,z}\quad
{^B}n_{2,z}\quad ]$ are excluded from the parameters vector $\vb*{\Phi}$, since
those three parameters are dependent on other parameters,
see \eqref{eq:10} and \eqref{eq:11}. That leaves us with $39-3=36$ parameters in
$\vb*{\Phi}$.

In this paper, we use a Denso VS060 6-\ac{dof} industrial manipulator with its
\textbf{DH} parameters presented in the first column of \tref{tab:params_comparison}. The Jacobian $\vb*{J}$ is computed by finite difference method. Applying the
identifiability analysis to the system, we found that there are 7 sets of
linearly dependent parameters out of the 36 parameters. These are exactly the same sets of parameters as found in \cite{Lembono2018}, although here we use two plane locations ($k=2$) instead of three.
For each set of the linearly dependent parameters, we can assign a fix value to one of the parameters. In this case, we fix the value of the parameters [$d_6, \theta_6, d_2, a_1, \alpha_1, \theta_1, d_1$] to their initial model's values. More 
details can be found in \cite{Lembono2018}.

Finally, we would like to note a few things:

\begin{itemize}

\item The reduction of the number of plane's locations from three to two do not
change the set of identifiable parameters (see~\cite{Lembono2018} for
the same analysis to three planes calibration). This validates our claim that
two plane's locations are sufficient for calibration using a 2D laser.

\item In this analysis, we only make use of the planar constraints to analyse
the identifiable parameters. This means that theoretically the planar
constraints are sufficient for the calibration purpose, although not practically
due to the various sources of errors (laser data error, flatness error, robot's
stiffness, etc), resulting in a slight dependency between the robot's and the
laser's parameters during optimization. Including the distance constraint in the
identifiability analysis will give us the same result in terms of the set of
identifiable parameters.

\item These results apply to most existing 6-\ac{dof} industrial robots whose
kinematic structures are similar to our Denso robot. Moreover, the analysis can be extended to any robot with arbitrary kinematic structures.

\end{itemize}

\section{Experimental Result}
\label{sec:experiment_result}

The experimental setup can be seen in \fref{fig:setup}. We use the
following equipment:

\begin{itemize}

\item DENSO VS060 6-\ac{dof} industrial robot arm.

\item Micro-Epsilon scanControl 2600-100 2D \ac{lrf}, attached on the robot's
flange. The \ac{lrf} has $100$~mm optimum range and $300$~mm maximum range, with
$0.012$~mm accuracy in its z-direction. The optimum range is used here for better
accuracy.

\item A custom tool, attached to the robot's flange. The diameter of the
tool-tip is less than $0.1$~mm.

\item A base plate, attached below the robot's base. The plate has holes
with $0.3$~mm diameter and $0.3$~mm depth.

\item An aluminium plate with flatness within $0.05$~mm.
There are a pair of holes with $0.3$~mm diameter and $0.3$~mm depth on the plate. The holes
are separated at $D~=~500$~mm distance with $0.01$~mm tolerance.
$D$ is chosen to be as large as possible such that small errors in robot's and laser's parameters result in large tool-tip distance errors.
\end{itemize}

In addition, we also use FARO Edge Arm as an additional validation tool for SCALAR. It can measure
XYZ positions within $0.05$~mm accuracy. Note that this tool is only used to further validate SCALAR, and it is not required in the actual calibration procedure.

The calibration procedure during the experiment is as follows. First, the
aluminium plate is placed at two different locations in the robot's workspace.
For each location, the robot is moved to $N=50$ different poses such that the
\ac{lrf}'s ray intersects the plate at each pose. The robot's joint angles and the laser data
are recorded. This gives us 100 poses, 60 of which will be used for
calibration (as proposed in \cite{Lembono2018}) while the remaining 40 poses are
used for validation. The value of $M$ is chosen as 40.

Next, the plate is moved to 50 different locations, and the robot is moved
such that the tool-tip touches the pair of holes on the plate consecutively
without changing orientation. The robot's joint angles are recorded. This gives
us 50 pair of joint angles, 15 of which will be used for calibration and 35 will
be used for validation.

Finally, the tool-tip position is calibrated using the hole at the base plate,
as explained in \sref{sec:tool-tip_calib}.

The following subsections will be as follows. \sref{sec:validation_method}
explains the three methods for validating the calibration result. In \sref{sec:determining_weight} the optimal value for $w$ is determined experimentally. \sref{sec:general_result} presents the general result obtained by SCALAR. In \sref{sec:plane_error}, \sref{sec:tool-tip_error}, and \sref{sec:farro_data}, we
compare the performance of SCALAR against three other cases:

\begin{itemize}

\item Nominal DH. In this case, we run the whole calibration procedure, but we
only optimize the laser parameters and the plane parameters while keeping the DH
parameters to the nominal value, which is obtained from the manufacturer.

\item Noisy DH. Similar to the Nominal DH, but here we fix the noisy
(instead of nominal) DH parameters while optimizing all the other parameters.
The noisy DH parameters are generated by introducing random errors to the
nominal DH parameters within the range of $\pm 2$~mm for linear parameters and
$\pm 1$~degrees for angular parameters.

\item SCALAR$_\alpha$. In this case, we use only the planar constraint in the objective
function as in \cite{Lembono2018}.

\end{itemize}

In \sref{sec:tool-tip_calib_effect}, we analyse the effect of the tool
calibration error to the SCALAR calibration result. We claim that the tool
calibration does not need to be very accurate in order for SCALAR to yield a good
result. This means that we can even skip the tool calibration and just use the
nominal tool-tip parameters from the CAD model or some rough measurement.
Finally, in \sref{sec:discussion} we discuss some of the interesting points from
the experimental results.

\subsection{Validation Method}
\label{sec:validation_method}

In this paper, the \emph{planar error} is defined as the error of the laser
data points with respect to the plane, and can be calculated by the following
formula,
\begin{equation}
  \delta_{p} =  |{{^B}\vb*{n}_k}^T \cdot {^B}\vb*{p}_{ji} - {^B}l_k| \quad .
\end{equation}

The \emph{tool-tip distance error} is defined as the difference between the
distance travelled by the tool-tip when it touches the pair of holes according
to the \emph{DH} parameters and according to the known value ($D$), and it is
calculated by the following formula,
\begin{equation}
  \label{eq:tool-tip_error}
  \delta_{t} =  ||{^B}\vb*{t}_{l2} - {^B}\vb*{t}_{l1}| - D | \quad .
\end{equation}

\begin{figure}[t]
  \centering
  \includegraphics[width=0.6\linewidth]{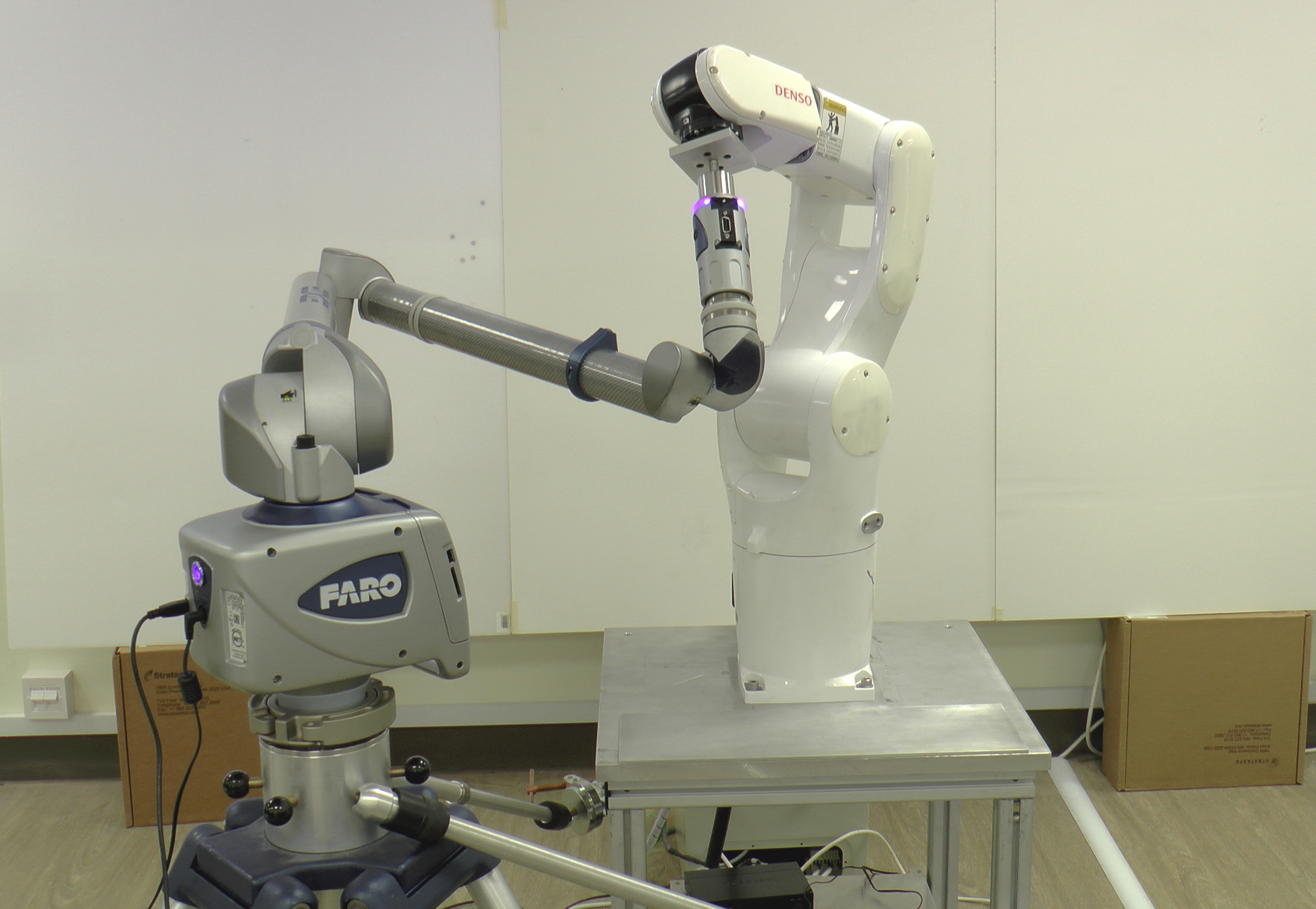} \quad
  \caption{FARO Edge Arm is used as the third method to validate the calibration
          result}
  \label{fig:faro}
\end{figure}

The planar error and the tool-tip distance errors will be used for validating
the calibration method.

The validation using the tool-tip distance error involves only the tool-tip and
the holes on the plate, so it does not require additional equipment. However,
the certainty provided by this validation method is limited by the size of the
holes and by how much the human eye can differentiate the tool-tip position. To
validate the calibration result with more certainty, we use a FARO Edge Arm as
the third validation method. The device can measure its end-point very
accurately within $0.05$~mm. The end-point is rigidly attached to the robot's
flange, as depicted in \fref{fig:faro}. The validation process here is similar
to the second method with the tool-tip; the robot is moved to two different
poses without changing the orientation. The distance between the two poses, $D$,
is measured by the FARO arm, and it is compared with the distance calculated based
on the DH parameters. \emph{FARO distance error} is defined as
\begin{equation}
  \label{eq:farro_error}
 \delta_{f} =  ||{^B}T_{E,2}{^E}\vb*{o} - {^B}T_{E,1}{^E}\vb*{o}| - D | \quad
\end{equation}
where ${^B}T_{E,2}$ and ${^B}T_{E,1}$ refer to the two poses of end-effector
with the same orientation and ${^E}\vb*{o}$ refers to the zero position vector
(the origin of the end-effector frame).

We use the mean, the standard deviation, and the maximum of $\delta_p$,
$\delta_t$, and $\delta_f$ to validate the proposed calibration method. For validation, we use
40 poses at two plate's locations for the calculation of $\delta_p$, 35 plate's
locations for $\delta_t$ and 50 pair of robot poses for $\delta_f$.

\renewcommand{\arraystretch}{0.9}
\begin{table}[t]
\caption{Comparison of the parameters: Nominal, Noisy, SCALAR$_\alpha$ and SCALAR}
\label{tab:params_comparison}
\centering
\begin{tabular}{c c c c c}
\toprule
\textbf{Parameter} & \textbf{Nominal} &  \textbf{Noisy}  & \textbf{SCALAR$_\alpha$}& \textbf{SCALAR+}\\
\midrule
${\alpha}_1$	&	0.00	&	0.00	&	0.00	&	0.00	\\
$a_1$			&	0.00	&	0.00	&	0.00	&	0.00	\\
${\theta}_1$	&	0.00	&	0.00	&	0.00	&	0.00	\\
$d_1$			&	345.00	&	345.00	&	345.00	&	345.00	\\
\midrule
${\alpha}_2$	&	-90.00	&	-90.17	&	-89.94	&	-89.92	\\
$a_2$			&	0.00	&	-2.00	&	-0.05	&	0.17	\\
${\theta}_2$	&	-90.00	&	-89.56	&	-90.01	&	-89.97	\\
$d_2$			&	0.00	&	-0.79	&	-0.79	&	-0.79	\\
\midrule
${\alpha}_3$	&	0.00	&	-0.71	&	0.01	&	0.01	\\
$a_3$			&	305.00	&	303.75	&	305.42	&	305.07	\\
${\theta}_3$	&	90.00	&	89.18	&	90.01	&	90.01	\\
$d_3$			&	0.00	&	-0.62	&	1.04	&	1.01	\\
\midrule
${\alpha}_4$	&	90.00	&	89.79	&	89.95	&	89.97	\\
$a_4$			&	-10.00	&	-10.32	&	-9.82	&	-9.83	\\
${\theta}_4$	&	0.00	&	-1.57	&	-0.15	&	-0.15	\\
$d_4$			&	300.00	&	300.74	&	300.05	&	299.61	\\
\midrule
${\alpha}_5$	&	-90.00	&	-90.59	&	-89.96	&	-89.97	\\
$a_5$			&	0.00	&	-1.89	&	0.25	&	0.22	\\
${\theta}_5$	&	0.00	&	0.76	&	0.04	&	0.03	\\
$d_5$			&	0.00	&	0.68	&	-0.37	&	-0.24	\\
\midrule
${\alpha}_6$	&	90.00	&	90.00	&	89.94	&	89.96	\\
$a_6$			&	0.00	&	0.00	&	0.04	&	0.08	\\
${\theta}_6$	&	0.00	&	0.00	&	0.00	&	0.00	\\
$d_6$			&	70.00	&	70.00	&	70.00	&	70.00	\\
\midrule
$p_x$	&	-127.50	&	-127.50	&	-125.46	&	-125.26	\\
$p_y$	&	-33.00	&	-33.00	&	-33.92	&	-34.04	\\
$p_z$	&	101.50	&	101.50	&	99.35	&	99.01	\\
$r_x$	&	0.00	&	0.00	&	-0.00	&	-0.00	\\
$r_y$	&	0.00	&	0.00	&	-0.00	&	-0.00	\\
$r_z$	&	1.00	&	1.00	&	1.00	&	1.00	\\
$r_w$	&	3.14	&	3.14	&	3.13	&	3.13	\\
\midrule
$n_{1,x}$	&	0.00	&	0.00	&	-0.01	&	-0.01	\\
$n_{1,y}$	&	0.00	&	0.00	&	-0.04	&	-0.04	\\
$n_{1,z}$	&	1.00	&	1.00	&	1.00	&	1.00	\\
$l_{1}$		&	0.00	&	0.00	&	-83.97	&	-84.17	\\
\midrule
$n_{2,x}$	&	0.00	&	0.00	&	-0.01	&	-0.01	\\
$n_{2,y}$	&	1.00	&	1.00	&	1.00	&	1.00	\\
$n_{2,z}$	&	0.00	&	0.00	&	0.00	&	0.00	\\
$l_{2}$		&	600.00	&	600.00	&	819.79	&	818.91	\\
\bottomrule
\end{tabular}
\end{table}

\begin{figure}[t]
  \centering
  \subfloat[]{\includegraphics[width=0.45\linewidth]{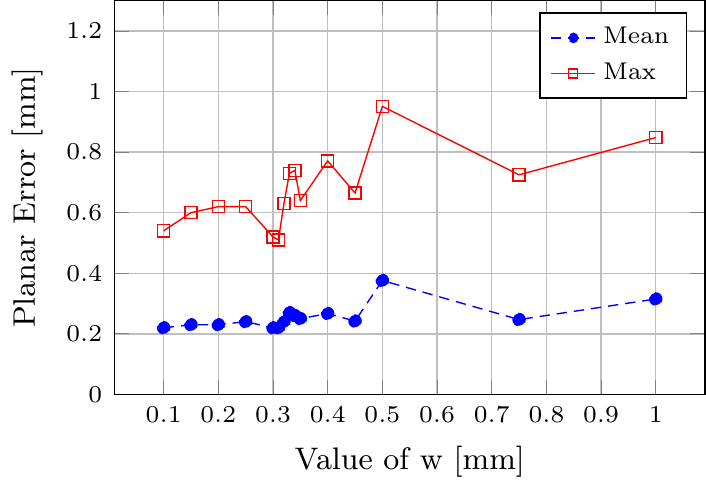}%
        \label{subfig:plane}}
  \subfloat[]{\includegraphics[width=0.45\linewidth]{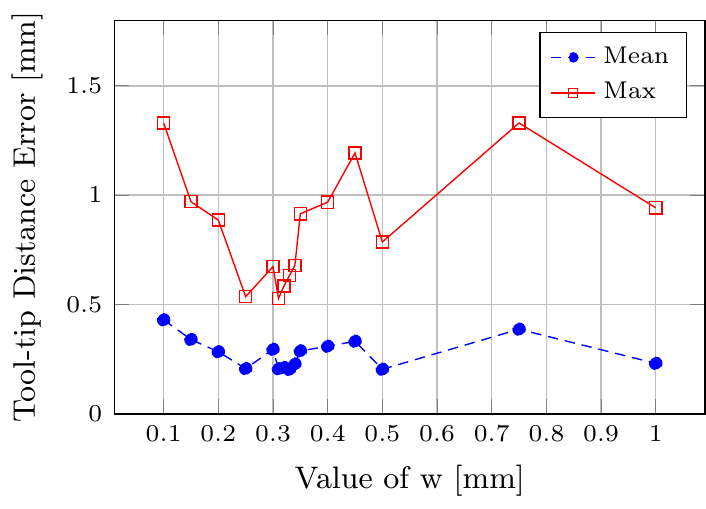}%
        \label{subfig:point}}
 \captionsetup{justification=centering}
  \caption{The effect of the weight $w$ on: a) Planar Error, and b) Tool-tip Distance Error}
  \label{fig:determine_w}
\end{figure}

\renewcommand{\arraystretch}{0.9}
\begin{table*}[th]
\caption{Comparing SCALAR Result with the other cases (in mm)}
\label{tab:comparing_scalar+}
\centering
\begin{tabular}{c c c c c c c c c c }
\toprule
\multirow{2}{*}{\textbf{Parameter} } & \multicolumn{3}{c}{\textbf{Planar Errors}} & \multicolumn{3}{c}{\textbf{Tool-tip Distance Errors}} & \multicolumn{3}{c}{\textbf{Faro Distance Errors}}\\
\cmidrule(lr){2-4}
\cmidrule(lr){5-7}
\cmidrule(lr){8-10}
& \textbf{Mean} &  \textbf{Std}  & \textbf{Max} & \textbf{Mean} &  \textbf{Std}  & \textbf{Max} & \textbf{Mean} &  \textbf{Std}  & \textbf{Max} \\
\midrule
Nominal DH \quad & 0.53 & 0.58 & 1.13 \quad & 0.741 & 0.736 & 1.424 \quad & 0.44 & 0.34 & 1.4\\
Noisy DH & 3.92 & 3.96 & 7.76 & 3.25 & 3.73 & 7.51 & 2.67 & 1.70 & 8.97\\
SCALAR$_\alpha$ & 0.20 & 0.22 & 0.46 & 0.664 & 0.608 & 1.338 & 0.34 & 0.26 & 1.45\\
SCALAR & 0.23 & 0.25 & 0.59 & 0.24 & 0.27 & 0.50 & 0.19 & 0.13 & 0.50\\
\bottomrule
\end{tabular}
\end{table*}

\subsection{Determining the weight $w$}
\label{sec:determining_weight}
The weight $w$ in the SCALAR objective function \eqref{eq:SCALAR+} determines the relative effect of the additional distance constraints term in the optimization. If $w$ is too small, the distance term will be dominated by the planar term, and the objective function will be reduced to the SCALAR$_\alpha$ objective function \eqref{eq:SCALAR}. If $w$ is too large, the distance term will dominate and the optimization result will be sub-optimal. To determine the optimal value of $w$,
the optimization is run while varying the value of $w$, and the result is compared in terms of the planar error and the tool-tip distance error.
\sfref{fig:determine_w}{subfig:plane} and \sfref{fig:determine_w}{subfig:point} show the effect of $w$ to the planar error and the tool-tip distance error, respectively. Note that when the value of $w$ is small, the planar error is small but the tool-tip distance error is large (this corresponds to SCALAR$_\alpha$ result). As $w$ is increased, the planar error increases slightly but the tool-tip distance error decreases, with its lowest point around $w$ = 0.31. As $w$ is increased further, both the planar error and the tool-tip distance error increase. Hence, $w$ = 0.31 is chosen as the optimal weight for all subsequent experiments.

\subsection{General Result}
\label{sec:general_result}

\tref{tab:params_comparison} shows the comparison between the nominal
parameters, the noisy parameters, and the calibrated parameters from SCALAR$_\alpha$
and SCALAR. The noisy parameters are used as the
initial values for SCALAR$_\alpha$ and SCALAR. It can be seen from the table that
although the noisy \emph{DH} parameters are far from the nominal ones,
SCALAR$_\alpha$ and SCALAR calibrate the \emph{DH} parameters to be very close
to the nominal ones. This is important because the nominal \emph{DH}
Parameters are already quite accurate (as will be shown in the following
sections). On the contrary, the calibrated plane parameters are very different
from the nominal and noisy ones, which demonstrate that accurate initial
estimate of the plane parameters are not required in SCALAR.

After calibration, SCALAR results in $0.23$~mm mean planar error, $0.24$~mm mean tool-tip distance error, and $0.19$~mm mean FARO distance error, which are better than the Nominal \emph{DH} and SCALAR$\alpha$ case. The optimization converges in less than 15 iterations with the total residual (Eq. \ref{eq:SCALAR+}) less than $10^{-6}~m^2$. 

Hayati et al. \cite{Hayati1985} established that in the case of
two parallel consecutive joints, the \emph{DH} parameters are singular w.r.t.
calibration. This means that a slight shift from the parallel configuration in
the actual joint location results in large changes to the singular \emph{DH}
parameters. The second and third joint of the robot are parallel in our case, so
this case should be considered as singular. Several ways have been proposed to
account for this singularity, including \cite{Hayati1985}. However, we realized
that by fixing the parameter $d_2$ (as a result of the identifiability analysis
in \sref{sec:third_step}), the singularity problem actually disappears. Fixing
$d_2$ corresponds to forcing the location of the second joint to be near the
original location, preventing the other set of \emph{DH} parameters to change
significantly. We also tried using the modified \emph{DH} parameters \cite{Hayati1985},
and the results that we obtained are similar to the ones presented here with
standard \emph{DH} parameters.

\subsection{Comparison of Planar Errors}
\label{sec:plane_error}
The first three columns of \tref{tab:comparing_scalar+} show the the planar error comparison.
While Noisy DH gives very large errors, SCALAR$_\alpha$ and SCALAR improve the
parameters such that the errors are even better than the Nominal DH. This
demonstrates that both methods are able to improve the parameters of the robot
and the laser as a whole such that the prediction of the laser data position can
be obtained accurately. SCALAR$_\alpha$ has slightly better result as compared to
SCALAR, which is expected since the only objective function in SCALAR$_\alpha$ is
the planar error.

\subsection{Comparison of Tool-tip Distance Errors}
\label{sec:tool-tip_error}

The middle columns of \tref{tab:comparing_scalar+} show the the tool-tip distance
error comparison. SCALAR has the lowest errors, while SCALAR$_\alpha$ has similar errors compared
to the Nominal DH. To explain this, note that the tool-tip distance error only
depends on the robot's kinematic parameters and not on the laser parameters.
This demonstrates that although SCALAR can optimize the combined robot kinematic
parameters and laser parameters to achieve good estimates for the laser data
points w.r.t. the robot's base frame, the robot kinematic parameters and the laser's extrinsic parameters, independently, are not optimal.
In other words, SCALAR$_\alpha$ modifies the kinematic parameters and the
laser parameters such that the combination minimizes the planar error, but
the kinematic parameters by itself do not predict properly the robot's
end-effector pose. By adding the distance constraint to the objective function
in SCALAR, the end-effector pose is predicted more accurately.

\subsection{Comparison of FARO Distance Error}
\label{sec:farro_data}

To further validate the calibration result, we use the FARO arm to measure the
distance $D$ travelled by the robot's end-effector. In contrast to
\sref{sec:tool-tip_error}, here the distance $D$ is not constant. The measured
distance is then compared with the calculated distance based on the DH
parameters, which gives us the FARO distance error. The comparison result is
shown in the last three columns of \tref{tab:comparing_scalar+}. The noisy DH parameters give very high errors
w.r.t. the FARO measurement, with $8.97$~mm as the maximum error. The mean
error of the nominal DH parameters is low, around $0.44$~mm, but the maximum
error is up to $1.41$~mm. SCALAR$_\alpha$ only improves the mean and standard deviation of the
error to be slightly better as compared to the nominal parameters. In contrast, SCALAR manages to improve the mean,
standard deviation, and maximum error significantly.

Here we emphasize that the conclusions obtained by looking at
FARO distance error and at tool-tip distance error are very similar: SCALAR
outperforms SCALAR$_\alpha$ in improving the robot's end-effector pose prediction. This
shows that at the absence of expensive equipment such as FARO arm, the
calibration and its validation can still be performed with good confidence.

\begin{figure}[t]
  \centering
  \includegraphics[height=40mm]{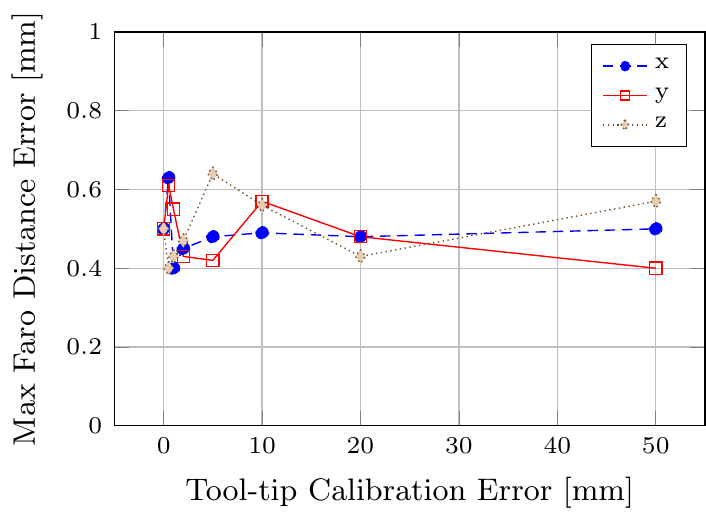}
   \caption{The effect of the tool-tip calibration error to the distance error}
   \label{fig:tip-calib-error}
\end{figure}

\subsection{Effect of the tool-tip calibration error}
\label{sec:tool-tip_calib_effect}

In this section, we demonstrate that accurate tool-tip calibration
is not necessary for SCALAR to successfully improve the kinematic and laser
parameters. To do that, we consider the calibrated tool-tip coordinates, see
\sref{sec:third_step}, to be the true tool-tip coordinates, and we introduce
errors separately to the x, y, and z coordinates of the tool-tip, ranging from
$0.5$~mm to $50$~mm. The modified tool-tip coordinates are used in SCALAR for
calibration. The calibrated kinematic parameters are then validated against the
FARO measurement by calculating the FARO distance errors such as in
\sref{sec:farro_data}.

\fref{fig:tip-calib-error} show the effect of the tool-tip coordinates
error in x, y, and z coordinate to the FARO distance error. It is clear that the
errors in tool-tip coordinates do not affect the FARO distance error
significantly. This is due to the fact that when the robot is moved to touch the
pair of holes on the plate, we keep the orientation of the robot constant, at
least according to the initial \emph{DH} parameters. When the orientation is
constant, the distance travelled by the robot flange is the same as the distance
travelled by the tool-tip, or any point which is rigidly attached to the robot's
flange. This means that the tool-tip coordinates do not
affect the tool-tip distance error at all.

However, it is important to note that in practice when we move the
robot to touch the pair of holes, we usually depend on the
robot's teach pendant to maintain the constant orientation of the robot. Since
the robot's teach pendant uses the nominal \emph{DH} parameters which have some errors,
the actual orientation may actually change between the two
robot poses, even though the teach pendant shows the two orientations to be the same.
If the actual change in orientation is large, the tool-tip coordinates errors can
have a significant effect on the calibration result.
The change in orientation will be propagated by the tool-tip coordinate error to result in the
tool-tip position error. In such cases, tool-tip calibration is necessary.

To evaluate the effect of orientation errors on the tool-tip
position, we can use a simple trigonometry equation which relates the magnitude
of an angle to the length of an arc,
\begin{equation}
  \label{eq:tool-tip_effect}
  {\epsilon}_x = {\epsilon}_{\theta} \cdot {\epsilon}_t \quad,
\end{equation}
where ${\epsilon}_x$ is the change in tool-tip position, ${\epsilon}_{\theta}$
is the change in tool-tip orientation, and ${\epsilon}_t$ is the error in
tool-tip coordinates. From this formula, given the initial robot orientation accuracy and
the acceptable error in tool-tip position, we can calculate the tolerable
tool-tip coordinate errors. For example, if the negligible
change in tool-tip position is $0.05$~mm, and the error in robot orientation
is $0.02$~rad ($1.15$~degrees), the acceptable error in tool-tip coordinates is
$2.5$~mm. This means that $2.5$~mm errors in tool-tip coordinates will only
result in $0.05$~mm error of the tool-tip position for a robot with orientation
accuracy around $1$~degree.

\subsection{Discussion}
\label{sec:discussion}

Based on the experiments that we conducted, we note the
followings:

\begin{itemize}

\item The laser accuracy depends on the plate's surface. Our plate
is made of aluminium, which gives quite a lot of noise to the laser data (within
the range of $0.1$~mm). If the plate can be made of other materials with less
specularity, the calibration result can be further improved.

\item The validation using the holes on the plate give us similar result to the
validation using FARO Edge Arm. This means that in
practice we can rely on the validation using the holes with confidence.

\item The addition of tool-tip errors to the objective function indeed results
in more inconvenience, as the data collection for this step requires manual
intervention. However, only 15 pair of points are required for calibration, so the manual step
can be completed quite fast (typically within 10-15 minutes). 
In addition, the distance constraint is not necessary if we only want to calibrate one of the two parameters set (either the robot's the laser's parameters) 

\item Instead of using 2D \ac{lrf}, it is possible to use two or more
pieces of 1D \ac{lrf} which are rigidly attached to the robot. This will cost
even cheaper as compared to 2D \ac{lrf}, although more calibration poses
might be necessary.

\item We demonstrate that using two plate's locations instead of three is
sufficient. The calibration algorithm can use more plate's locations so that the range of the workspace explored by the calibration is extended, but we found that the result using more plate's
locations is similar to that obtained by two plate's locations.

\item In this work we do not try to find the optimal set of calibration poses, 
but instead rely on the random poses to cover the robot's configuration space. There are a lot of methods that have been proposed to find such an optimal set, e.g. using the singular values of the Jacobian or the information matrix (~\cite{Hollerbach1996},\cite{xiong2017product}). \cite{rebello2017autonomous} use the parameters covariance matrix to determine the next-best-pose. 

\end{itemize}

\section{Conclusions}
\label{sec:conclusions}

In this paper, we have proposed SCALAR, a novel method to calibrate
simultaneously a 6-\ac{dof} industrial robot's kinematic parameters and a 2D
\ac{lrf} extrinsic parameters. SCALAR has been shown to be an improvement over our previous method (SCALAR$_\alpha$), since it finds the optimal parameters for each set
(robot and laser). SCALAR reduces the mean planar errors from $0.53$~mm to around
$0.23$~mm, the mean tool-tip distance errors from $0.74$~mm to $0.24$~mm, and the FARO distance errors from $0.44$~mm to $0.19$~mm. The proposed method is inexpensive and
convenient, as it does not require expensive external measurement systems or an
elaborated calibration setting.

\section*{Acknowledgment}
This work was supported by SMART Innovation Grant NG000074-ENG, NTUitive Gap Fund NGF-2016-01-028, by National Research Foundation, Prime Minister’s Office, Singapore, under its Medium-Sized Centre funding scheme, and by MEMMO project (Memory of Motion, \url{http://www.memmo-project.eu/}), funded by the European Commission's Horizon 2020 Programme (H2020/2018-20) under grant agreement 780684.

\bibliographystyle{IEEEtran}
\bibliography{references}

\begin{thebibliography}{10}
\providecommand{\url}[1]{#1}
\csname url@samestyle\endcsname
\providecommand{\newblock}{\relax}
\providecommand{\bibinfo}[2]{#2}
\providecommand{\BIBentrySTDinterwordspacing}{\spaceskip=0pt\relax}
\providecommand{\BIBentryALTinterwordstretchfactor}{4}
\providecommand{\BIBentryALTinterwordspacing}{\spaceskip=\fontdimen2\font plus
\BIBentryALTinterwordstretchfactor\fontdimen3\font minus
  \fontdimen4\font\relax}
\providecommand{\BIBforeignlanguage}[2]{{%
\expandafter\ifx\csname l@#1\endcsname\relax
\typeout{** WARNING: IEEEtran.bst: No hyphenation pattern has been}%
\typeout{** loaded for the language `#1'. Using the pattern for}%
\typeout{** the default language instead.}%
\else
\language=\csname l@#1\endcsname
\fi
#2}}
\providecommand{\BIBdecl}{\relax}
\BIBdecl

\bibitem{Zhang2018}
X.~Zhang, M.~Li, J.~H. Lim, Y.~Weng, Y.~W.~D. Tay, H.~Pham, and Q.-C. Pham,
  ``Large-scale 3d printing by a team of mobile robots,'' \emph{Automation in
  Construction}, vol.~95, pp. 98--106, 2018.

\bibitem{Suarez-Ruiz2018}
F.~Su{\'{a}}rez-Ruiz, T.~{Santoso Lembono}, and Q.-C. Pham, ``{RoboTSP – A
  Fast Solution to the Robotic Task Sequencing Problem},'' in \emph{IEEE Int.
  Conf. on Robotics and Automation}, 2018.

\bibitem{Lembono2018}
T.~S. Lembono, F.~Su{\'{a}}rez-Ruiz, and Q.~C. Pham, ``{SCALAR - Simultaneous
  Calibration of 2D Laser and Robot's Kinematic Parameters Using Three Planar
  Constraints},'' in \emph{IEEE/RSJ Int. Conf. on Intelligent Robots and
  Systems}, 2018.

\bibitem{Ye2006}
S.~H. Ye, Y.~Wang, Y.~J. Ren, and D.~K. Li, ``{Robot calibration using
  iteration and differential kinematics},'' \emph{Journal of Physics:
  Conference Series}, vol.~48, no.~1, pp. 1--6, 2006.

\bibitem{Ginani2011}
L.~S. Ginani and J.~M. S.~T. Motta, ``{Theoretical and practical aspects of
  robot calibration with experimental verification},'' \emph{Journal of the
  Brazilian Society of Mechanical Sciences and Engineering}, vol.~33, no.~1,
  pp. 15--21, 2011.

\bibitem{Nubiola2013}
A.~Nubiola and I.~A. Bonev, ``{Absolute calibration of an ABB IRB 1600 robot
  using a laser tracker},'' \emph{Robotics and Computer-Integrated
  Manufacturing}, vol.~29, no.~1, pp. 236--245, 2013.

\bibitem{Wu2017}
L.~Wu and H.~Ren, ``{Finding the Kinematic Base Frame of a Robot by Hand-Eye
  Calibration Using 3D Position Data},'' \emph{IEEE Transactions on Automation
  Science and Engineering}, vol.~14, no.~1, pp. 314--324, 2017.

\bibitem{Gatla2007}
C.~S. Gatla, R.~Lumia, J.~Wood, and G.~Starr, ``{An automated method to
  calibrate industrial robots using a virtual closed kinematic chain},''
  \emph{IEEE Transactions on Robotics}, vol.~23, no.~6, pp. 1105--1116, 2007.

\bibitem{Chiu2003}
Y.-J. Chiu and M.-H. Perng, ``{Self-calibration of a general hexapod
  manipulator using cylinder constraints},'' \emph{International Journal of
  Machine Tools and Manufacture}, vol.~43, pp. 1051--1066, 2003.

\bibitem{Zhong1995}
X.~L. Zhong and J.~M. Lewis, ``{New method for autonomous robot calibration},''
  \emph{IEEE Int. Conf. on Robotics and Automation}, vol.~2, pp. 1790--1795,
  1995.

\bibitem{Zhong1996}
X.~L. Zhong, J.~M. Lewis, and F.~L.~N. Nagy, ``{Autonomous robot calibration
  using a trigger probe},'' \emph{Robotics and Autonomous Systems}, vol.~18,
  no.~4, pp. 395--410, 1996.

\bibitem{Ikits1997}
M.~Ikits and J.~Hollerbach, ``{Kinematic calibration using a plane
  constraint},'' \emph{IEEE Int. Conf. on Robotics and Automation}, vol.~2,
  no.~4, pp. 3191--3196, 1997.

\bibitem{Zhuang1999}
Z.~Hanqi, S.~Motaghedi, and Z.~Roth, ``{Robot calibration with planar
  constraints},'' \emph{IEEE Int. Conf. on Robotics and Automation}, vol.~1,
  pp. 1--25, 1999.

\bibitem{Zilong2006}
T.~Zilong, N.~Zhanhai, and L.~Xinggang, ``{Autonomous calibration research of
  polishing robot},'' \emph{Proceedings of the World Congress on Intelligent
  Control and Automation (WCICA)}, vol.~2, pp. 8938--8942, 2006.

\bibitem{Hage2011}
H.~Hage, P.~Bidaud, and N.~Jardin, ``{Practical consideration on the
  identification of the kinematic parameters of the St¨aubli TX90 robot},''
  \emph{World}, pp. 19--25, 2011.

\bibitem{Joubair2015}
A.~Joubair and I.~A. Bonev, ``{Non-kinematic calibration of a six-axis serial
  robot using planar constraints},'' \emph{Precision Engineering}, vol.~40, pp.
  325--333, 2015.

\bibitem{Wang2009}
W.~Wang, A.~Li, and D.~Wu, ``Robot calibration by observing a virtual fixed
  point,'' in \emph{IEEE International Conference on Robotics and Biomimetics
  (ROBIO)}, Dec 2009, pp. 1351--1355.

\bibitem{Liu2009}
Y.~Liu, N.~Xi, G.~Zhang, X.~Li, H.~Chen, C.~Zhang, M.~J. Jeffery, and T.~A.
  Fuhlbrigge, ``An automated method to calibrate industrial robot joint offset
  using virtual line-based single-point constraint approach,'' in
  \emph{IEEE/RSJ Int. Conf. on Intelligent Robots and Systems}, 2009, pp.
  715--720.

\bibitem{Liu2011low}
Y.~Liu and N.~Xi, ``Low-cost and automated calibration method for joint offset
  of industrial robot using single-point constraint,'' \emph{Industrial Robot},
  vol.~38, no.~6, pp. 577--584, 2011.

\bibitem{Joubair2015kinematic}
A.~Joubair and I.~A. Bonev, ``Kinematic calibration of a six-axis serial robot
  using distance and sphere constraints,'' \emph{The International Journal of
  Advanced Manufacturing Technology}, vol.~77, no. 1-4, pp. 515--523, 2015.

\bibitem{Choi2018}
C.~L. Choi, J.~Rebello, L.~Koppel, P.~Ganti, A.~Das, and S.~L. Waslander,
  ``Encoderless gimbal calibration of dynamic multi-camera clusters,'' in
  \emph{IEEE Int. Conf. on Robotics and Automation}, May 2018, pp. 2126--2133.

\bibitem{ArunDas2018}
A.~Das, ``Informed data selection for dynamic multi-camera clusters,'' Ph.D.
  dissertation, University of Waterloo, 2018.

\bibitem{Zhang2004}
{Qilong Zhang} and R.~Pless, ``{Extrinsic calibration of a camera and laser
  range finder (improves camera calibration)},'' in \emph{IEEE/RSJ Int. Conf.
  on Intelligent Robots and Systems}, vol.~3, 2004, pp. 2301--2306.

\bibitem{Unnikrishnan2005}
R.~Unnikrishnan and M.~Hebert, ``{Fast Extrinsic Calibration of a Laser
  Rangefinder to a Camera},'' \emph{Robotics}, 2005.

\bibitem{Chen2018}
W.~Chen, J.~Du, W.~Xiong, Y.~Wang, S.~Chia, B.~Liu, J.~Cheng, and Y.~Gu, ``{A
  Noise-Tolerant Algorithm for Robot-Sensor Calibration Using a Planar Disk of
  Arbitrary 3-D Orientation},'' \emph{IEEE Transactions on Automation Science
  and Engineering}, vol.~15, no.~1, pp. 251--263, 2018.

\bibitem{Li2017}
Y.~Li, W.~Zhang, J.~Cao, J.~Peng, B.~Jiang, and L.~Wang, ``{3D laser scanner
  calibration method based on invasive weed optimization and
  Levenberg-Marquardt algorithm},'' in \emph{IEEE Int. Conf. on Automation
  Science and Engineering}, 2017, pp. 1280--1285.

\bibitem{Newville2014}
M.~Newville, T.~Stensitzki, D.~B. Allen, and A.~Ingargiola, ``{LMFIT:
  Non-Linear Least-Square Minimization and Curve-Fitting for Python},'' 2014.

\bibitem{Hollerbach1996}
J.~M. Hollerbach and C.~W. Wampler, ``{The calibration index and taxonomy for
  robot kinematic calibration methods},'' \emph{International Journal of
  Robotics Research}, vol.~15, no.~6, pp. 573--591, 1996.

\bibitem{Hayati1985}
S.~Hayati and M.~Mirmirani, ``{Improving the absolute positioning accuracy of
  robot manipulators},'' \emph{Journal of Robotic Systems}, vol.~2, no.~4, pp.
  397--413, 1985.

\bibitem{xiong2017product}
G.~Xiong, Y.~Ding, L.~Zhu, and C.-Y. Su, ``A product-of-exponential-based robot
  calibration method with optimal measurement configurations,''
  \emph{International Journal of Advanced Robotic Systems}, vol.~14, no.~6,
  2017.

\bibitem{rebello2017autonomous}
J.~Rebello, A.~Das, and S.~Waslander, ``Autonomous active calibration of a
  dynamic camera cluster using next-best-view,'' in \emph{IEEE/RSJ Int. Conf.
  on Intelligent Robots and Systems}, 2017, pp. 1484--1489.

\end{thebibliography}

\end{document}